\documentclass[10pt,twocolumn,letterpaper]{article}
\usepackage{iccv}
\usepackage{times}
\usepackage{epsfig}
\usepackage{graphicx}
\usepackage{amsmath}
\usepackage{amssymb}
\iccvfinalcopy

\ificcvfinal\fi
\usepackage{color,xcolor}

\usepackage{amsmath}
\usepackage{bm}

\usepackage{algpseudocode}
\usepackage{amssymb}
\usepackage{booktabs}
\usepackage{array}
\usepackage{makecell}
\usepackage{arydshln}
\usepackage{multirow}
\usepackage{threeparttable} 
\usepackage[pagebackref=true,breaklinks=true,letterpaper=true,colorlinks,bookmarks=false,citecolor=blue,linkcolor=red]{hyperref}

\newcommand\bb[1]{\textbf{#1}}

\newcommand\ul[1]{\underline{#1}}

\usepackage{pifont}
\newcommand{\cmark}{\ding{51}}
\newcommand{\xmark}{\ding{55}}

\begin{document}
\title{Dynamical Isometry: The Missing Ingredient for Neural Network Pruning}

\author{Huan Wang, Can Qin, Yue Bai, Yun Fu \\
Northeastern University, Boston, MA, USA \\
{\tt\small {huan.wang.cool@gmail.com, \{qin.ca, bai.yue\}@northeastern.edu, yunfu@ece.neu.edu}}
}

\maketitle
\ificcvfinal\thispagestyle{empty}\fi

\begin{abstract}
Several recent works~\cite{renda2020comparing,le2021network} observed an interesting phenomenon in neural network pruning: A larger finetuning learning rate can improve the final performance significantly. Unfortunately, the reason behind it remains elusive up to date. This paper is meant to explain it through the lens of dynamical isometry~\cite{saxe2014exact}. Specifically, we examine neural network pruning from an unusual perspective: pruning as initialization for finetuning, and ask \emph{whether the inherited weights serve as a good initialization for the finetuning?} The insights from dynamical isometry suggest a negative answer. Despite its critical role, this issue has not been well-recognized by the community so far. In this paper, we will show the understanding of this problem is very important -- on top of explaining the aforementioned mystery about the larger finetuning rate, it also unveils the mystery about the value of pruning~\cite{crowley2018closer,liu2019rethinking}. Besides a clearer theoretical understanding of pruning, resolving the problem can also bring us considerable performance benefits in practice.
\vspace{-2em}
\end{abstract}

\section{Introduction}
Pruning is a time-honored methodology to reduce parameters in a neural network without seriously compromising its performance~\cite{Ree93,sze2017efficient}. The prevailing pipeline of pruning comprises three steps: 1) pretraining: train a dense model; 2) pruning: prune the dense model based on certain rules; 3) finetuning: retrain the pruned model to regain performance. Most existing research focuses on the second step, which is believed to be the central problem in pruning, \ie, seeking the best criterion to select unimportant weights so as to incur as less performance degradation as possible. 

These said several recent works discovered that there are other important axes not well noted by the community. Specifically, \cite{renda2020comparing,le2021network} found that the learning rate (LR) in finetuning holds a critical role in the final performance. A proper learning rate schedule (\eg, a larger initial LR 0.01 vs.~0.001) can improve the top-1 accuracy of a pruned ResNet-34 model~\cite{resnet} by more than 1\% on ImageNet~\cite{imagenet}. This discovery calls upon more attention being paid to the finetuning step when comparing different pruning methods. Unfortunately, they did not present more theoretical insights to explain its occurrence. Up to date, this remains an open problem in the community, to our best knowledge. 

Another open problem in pruning is the debate about its value. Several works have questioned the necessity of the 1st step in the pipeline. \cite{crowley2018closer,liu2019rethinking} argue the 1st step (\ie, pretraining a large model) is \emph{unnecessary} because they empirically found the small model \emph{trained from scratch} can match (or sometimes outperform) the counterpart pruned from the pre-trained large model. Pruning has been developed for more than 30 years to date~\cite{mozer1989skeletonization,karnin1990simple,OBD}. Even a moment's thought is enough to see it \emph{should} have value. However, at the same time, the discovery of \cite{crowley2018closer,liu2019rethinking} is challenging our belief. To our best knowledge, no paper has systematically responded to the questioning and explained why scratch training could match pruning in their experiments.

\begin{figure}[t]
\centering
\begin{tabular}{c@{\hspace{0.005\linewidth}}c@{\hspace{0.005\linewidth}}c@{\hspace{0.005\linewidth}}c@{\hspace{0.005\linewidth}}c@{\hspace{0.005\linewidth}}c}
\includegraphics[width=\linewidth]{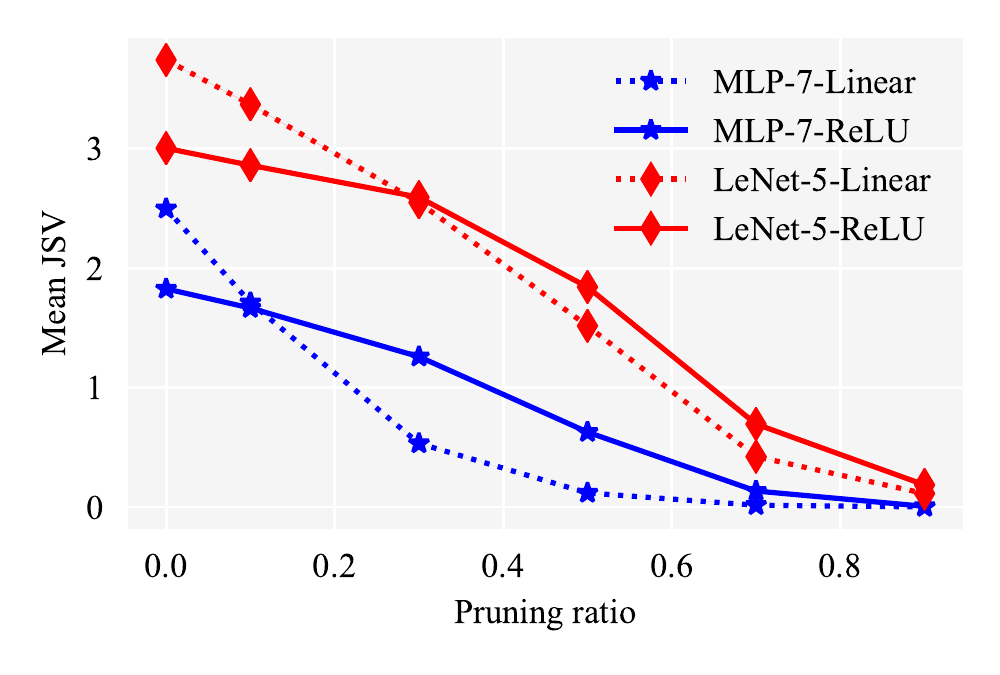} &
\end{tabular}
\caption{Dynamical isometry measured by mean JSV (Jacobian singular value) (defined in Eq.~(\ref{eq:mean_jsv})) of pruned networks w.r.t.~different pruning ratios on MNIST dataset. Pruning ratio 0 indicates the unpruned networks. Note, with a larger pruning ratio, mean JSV (dynamical isometry) is hurt more, implying that pruning actually serves as a \emph{poor} initialization for later finetuning.}
\label{fig:pruning_hurts_jsv}
\vspace{-1em}
\end{figure}

These are two open problems in neural network pruning, which are discovered empirically but with little theoretical understanding of the underlying rationale. On the other hand, the prevailing point of view about the pruning pipeline is to regard \emph{finetuning as a post-processing step for pruning}\footnote{As a result, many pruning papers describe the pruning part in length, leaving the details of finetuning scratched or even not mentioned at all. This issue was also noted by~\cite{le2021network}.}. Actually, there is another perspective around, which we think is better, is to see \emph{pruning as initialization for finetuning}. This perspective holds \emph{finetuning at the center} and asks if pruning serves it well. After all, the performance \emph{after} finetuning is what really matters; there is no reason to overlook it~\cite{le2021network}. From this perspective, it is easy to see a seriously under-explored problem in pruning: In stark contrast to the fact that the initialization of neural networks in SGD training~\cite{robbins1951stochastic,bottou2010large} has been well-recognized~\cite{glorot2010understanding,sutskever2013importance,mishkin2015all,krahenbuhl2015data,he2015delving}, we have rarely questioned \emph{if pruning provides the proper initialization for finetuning} (note finetuning is essentially SGD training, too). 

In this paper, we examine the initialization role of pruning from the signal propagation perspective of neural networks. Specifically, we look into how pruning and finetuning affect the \emph{dynamical isometry}~\cite{saxe2014exact} of a pretrained network. With its help, our results suggest theoretical explanations towards the two aforementioned mysteries in pruning: (1) Pruning hurts dynamical isometry (see Fig.~\ref{fig:pruning_hurts_jsv}). \emph{Finetuning can recover it}. A \emph{larger} LR can help recover it \emph{faster} thus achieving a better final accuracy. (2) The finetuning learning rates in \cite{crowley2018closer,liu2019rethinking} are actually \emph{far from optimal}, which cannot fully recover the dynamical isometry and finally lead to the ``no value of pruning'' argument. With a more proper finetuning LR, our results suggest that filter pruning (even if using the most basic $L_1$-norm pruning~\cite{li2017pruning}) still \emph{outperforms} scratch training across different pruning ratios.

On top of resolving the two mysteries, our work also suggests that \emph{dynamical isometry recovery} (DIR) before finetuning is a very imperative step, which should be included in the pruning pipeline. In this regard, we propose a DIR method (OrthP) based on inherited weights by pruning.

In short, our contributions in this paper can be summarized into the following three aspects:
\begin{itemize}
    \item We tap into dynamical isometry as a powerful tool to diagnose the role of pruning as initialization and discover that the inherited weights by pruning are \emph{not} a good initialization (Sec.~\ref{subsec:DI}).
    \item We propose a dynamical isometry recovery algorithm (OrthP) that can be applied in a plug-and-play fashion to recover the damaged dynamical isometry (Sec.~\ref{subsec:OrthP}).
    \item With the insights from dynamical isometry and OrthP, we unveil two mysteries in pruning: why a larger finetuning LR can improve the final performance significantly~\cite{renda2020comparing,le2021network} (Sec.~\ref{subsec:analysis_mlp}) and whether filter pruning really bears no value compared to training from scratch~\cite{crowley2018closer,liu2019rethinking} (Sec.~\ref{subsec:value_of_pruning}).
\end{itemize}

\section{Related Work} \label{sec:related_work}
\noindent \bb{Conventional pruning}. Pruning aims to remove as many parameters as possible in a neural network meanwhile maintaining its performance. There are many ways to categorize pruning methods. The most popular two are grouping by pruning structure and methodology. 

(1) In terms of structure, pruning can be specified into unstructured pruning~\cite{han2015learning,han2015deep} and structured pruning~\cite{wen2016learning,li2017pruning,he2017channel}. For the former, a single weight is the basic pruning element. Unstructured pruning can deliver a high compression ratio; whereas, without regularization, the pruned locations usually spread \emph{randomly} in the network, which is hard to exploit for acceleration. On the opposite, structured pruning introduces certain patterns in the pruned locations, which benefit subsequent acceleration while cannot achieve as much compression. Choices between unstructured and structured pruning depend on specific application needs. For structured pruning, there are still many sub-groups~\cite{mao2017exploring}. In the literature, without specific mention, structured pruning means filter pruning or channel pruning. 

(2) In terms of pruning methodology (\ie, how to select unimportant weights to prune), pruning falls into two paradigms in general: importance-based and penalty-based. The former prunes weights based on some established importance criteria, such as magnitude (for unstructured pruning)~\cite{han2015learning,han2015deep} or $L_1$-norm (for filter pruning)~\cite{li2017pruning}, saliency based on 2nd-order gradients (\eg, Hessian or Fisher)~\cite{OBD,OBS,theis2018faster,wang2019eigendamage,singh2020woodfisher}. The latter adds a penalty term to the objective function, drives unimportant weights towards zero, then removes those with the smallest magnitude. Note, the two groups are \emph{not} starkly separated. Many methods take wisdom from both sides. For example,~\cite{DinDinHanTan18,wang2018structured,wang2021neural} select unimportant weights by magnitude (akin to the first group) while also employing the regularization to penalize weights (akin to the second group). There is no conclusion about which paradigm is better, yet empirically, the state-of-the-art pruning methods are closer to the second paradigm, \ie, deciding weights via training instead of some derived formulas. Although no theories have formally discussed the reason, we can take a rough guess with the knowledge from this paper: Training can 
recover dynamical isometry, which is beneficial to subsequent finetuning.

For more comprehensive literature, we refer interested readers to several surveys: an outdated one~\cite{Ree93}, some recent surveys of pruning alone~\cite{gale2019state,blalock2020state} or pruning as a sub-topic under the general umbrella of model compression and acceleration~\cite{sze2017efficient,cheng2018recent,cheng2018model,deng2020model}.

\vspace{0.500em}
\noindent \bb{Pruning at initialization (PaI)}. Recent years have seen some new pruning paradigms. The most prominent one is pruning at initialization. Different from the conventional pruning, which prunes a \emph{pretrained} model, PaI methods prune a \emph{randomly initialized} model. Existing PaI approaches mainly include~\cite{lee2019snip,lee2020signal,wang2020picking,frankle2021pruning,ramanujan2020what} and the series of lottery ticket hypothesis~\cite{frankle2019lottery,frankle2020linear}. We refer interested readers to~\cite{wang2021emerging} for a more comprehensive summary about this new pruning paradigm. This topic is relevant to this work mainly because one PaI paper~\cite{lee2020signal} also examines pruning using the tool of dynamical isometry. Our work is different from theirs in that (1) they focus on pruning a random network while ours focus on the broader conventional pruning; (2) we show how dynamical isometry can be used to unveil two important open problems in pruning.

\section{Methodology and Exploration}
\subsection{Prerequisite: Dynamical Isometry} \label{subsec:DI}
Dynamical isometry (DI) is studied in the topic of trainability of deep neural networks. It was first brought up in~\cite{saxe2014exact}, where a deep linear network is analyzed. To extend it to convolutional networks,~\cite{xiao2018dynamical} proposes delta-orthogonalization and successfully trains a 10,000-layer vanilla CNN. Despite the promising progress, in terms of practical performance, these networks are far from state-of-the-art networks (such as ResNets~\cite{resnet}). 

Specifically, dynamical isometry is defined as \emph{the singular values of Jacobian matrix being around 1}~\cite{saxe2014exact}. It has been shown that very deep neural networks are trainable if they are initialized to meet dynamical isometry. For linear networks, dynamical isometry can be achieved \emph{exactly} by orthogonal initialization~\cite{saxe2014exact}; while for neural networks with non-linearity (like ReLU) and convolution, it can only be approximated up to date~(see Tab.~\ref{tab:jsv_different_nets}).

\vspace{0.500em}
\noindent \bb{Mean Jacobian singular values}. Since dynamical isometry is measured by the Jacobian singular values (JSV's), we adopt the \emph{mean of Jacobian singular values} (denoted by $\bar{S}$) as a scalar metric for analysis. Specifically, for a Jacobian  $\mathbf{J} \in \mathbb{R}^{C \times D_{\text{in}}}$ ($C$ stands for the output dimension, \ie, the number of classes, $D_{\text{in}}$ for the input dimension), apply singular value decomposition~\cite{trefethen1997numerical} to it,
\begin{equation}
U, \Sigma, V = \text{svd}(\mathbf{J}), \bar{S} = \frac{1}{K}\sum_{i=1}^{K} \Sigma_{ii},
\label{eq:mean_jsv}
\end{equation}
where $\Sigma$ is the singular value matrix and $K=\min(C, D_{\text{in}})$.

\vspace{0.5em}
\noindent \bb{Pruning as \emph{poor} initialization}. We investigate how the pruning affects $\bar{S}$ (note, we focus on \emph{structured pruning (\ie, filter pruning)} in this paper). The results are shown in Fig.~\ref{fig:pruning_hurts_jsv}. As seen, the mean JSV is consistently \emph{damaged} by pruning; and a larger pruning ratio, more decrease of the mean JSV. This means, pruning actually servers as a very \emph{poor} initialization scheme for the subsequent finetuning. 

In stark contrast to the broad awareness that initialization is rather critical to neural network training~\cite{glorot2010understanding,sutskever2013importance,mishkin2015all,krahenbuhl2015data,he2015delving}, the initialization role of pruning has received negligible research attention, however. As far as we know, no prior works have noted this issue when pruning a pretrained network or tried to recover the damaged dynamical isometry before finetuning. Next, we are going to show what the consequences are if this issue is not well-attended. The results will help us understand why finetuning LR is so important to the final pruning performance.

\begin{table}[t]
\centering  
\setlength\tabcolsep{4pt}
\begin{tabular}{lccc}
\toprule
Network         & JSV mean (std)        & JSV max       & JSV min \\
\midrule
MLP-7-Linear    & 1.0000 (0.0000)       & 1.0000        & 1.0000 \\
MLP-7-ReLU      & 1.2268 (0.5519)       & 3.2772        & 0.2282 \\
LeNet-5-Linear  & 0.9983 (0.0842)       & 1.2330        & 0.7896 \\
LeNet-5-ReLU    & 1.8331 (0.5731)       & 3.6007        & 0.6151 \\
\bottomrule
\end{tabular}
\vspace{0.5000em}
\caption{JSVs (Jacobian singular values) of orthogonal initialization~\cite{saxe2014exact} on different types of neural networks on MNIST dataset. Note, only the MLP linear network can achieve dynamical isometry \emph{exactly} (\ie, all the JSVs equal to 1).} 
\label{tab:jsv_different_nets}
\vspace{-1em}
\end{table}

\begin{table*}[t]
\centering  
\setlength\tabcolsep{10pt}
\begin{tabular}{|l|c|c|}
\hline
                        &       Initial LR 0.01                                             & Initial LR 0.001 \\
\hline
For Hypothesis 1        &       90 epochs, 0:0.01,30:0.001,60:0.0001                        & 90 epochs, 0:0.001,45:0.001  \\
For Hypothesis 2        &       900 epochs, 0:0.01,300:0.001,600:0.0001                     & 900 epochs, 0:0.001,450:0.001 \\
For Hypothesis 3        &       OrthP, 90 epochs, 0:0.01,30:0.001,60:0.0001                 & OrthP, 90 epochs, 0:0.001,45:0.001 \\
For Hypothesis 4        &       OrthP, 900 epochs, 0:0.01,300:0.001,600:0.0001              & OrthP, 900 epochs, 0:0.001,450:0.001 \\
\hline
\end{tabular}
\vspace{0.5000em}
\caption{Summary of different finetuning LR schedules (corresponding to the 4 proposed hypotheses in Sec.~\ref{subsec:analysis_mlp}). For reference, the unpruned model is trained with LR schedule ``90 epochs, 0:0.01,30:0.001,60:0.0001''. OrthP is the proposed orthogonalization scheme (Sec.~\ref{subsec:OrthP}). LR schedule ``90 epochs, 0:0.01,30:0.001,60:0.0001'' means that the total number of training epochs is 90; at epoch 0, LR is set to 0.01; then at epoch 30, it decays to 0.001; then at epoch 60, it decays to 0.0001. The others can be inferred accordingly.}
\label{tab:mlp_lr_schedule}
\vspace{-1em}
\end{table*}

\subsection{Dynamical Isometry Recovery in Filter Pruning} \label{subsec:OrthP}
In~\cite{saxe2014exact}, they propose a weight orthogonalization scheme to achieve dynamical isometry for neural network \emph{initialization}, namely, the initial weights are \emph{randomly sampled}. Different from their case, here the initial weights are inherited from a pretrained model by pruning. Therefore, we need to adapt it to our application. 

For a fully-connected layer parameterized by a matrix $W_0 \in \mathbb{R}^{J\times K}$ (for a convolutional layer parameterized by a 4-d tensor of shape $\mathbb{R}^{N\times C\times H\times W}$, it can be reshaped to a matrix of shape $\mathbb{R}^{N\times CHW}$), it reduces to matrix $W$ of size $\mathbb{R}^{J_1 \times M_1}$ ($J_1 \le J, K_1 \le K$) after structured pruning. Then, we apply the weight orthogonalization technique~\cite{mezzadri2006generate} based on QR decomposition~\cite{trefethen1997numerical} to $W$,
\begin{equation}
\begin{split}
    Q, R &= \text{qrd}(W), \\
    W^* &= Q \odot \text{sign}(\text{diag}(R)),
\end{split}
\label{eq:orth_linear}
\end{equation}
where qrd($\cdot$) stands for the QR decomposition; $Q$ is an orthogonal matrix of the same size as $W$ ($\mathbb{R}^{J_1\times K_1}$); $R$ is an an upper triangular matrix of size $\mathbb{R}^{K_1\times K_1}$; sign($\cdot$) is the sign function which returns the positive or negative sign of its argument; $\odot$ represents the Hadamard (element-wise) product aligned to the last axis (since $Q$ and \text{sign}(\text{diag}(R)) share the same dimension at the last axis). 

As an orthogonalized version of $W$, $W^*$ recovers the dynamical isometry damaged by pruning. Therefore, we propose to employ $W^*$ instead of the original $W$ as the initialization weights for later finetuning. We dub this weight orthogonalization method for \emph{pruned} models as \bb{OrthP}.

With this tool, in the next two subsections, we present our investigation path with empirical results to show how dynamical isometry can be used to unveil the two mysteries in pruning: Sec.~\ref{subsec:analysis_mlp} answers why a larger finetuning LR can improve the final performance in pruning~\cite{renda2020comparing,le2021network}; Sec.~\ref{subsec:value_of_pruning} discusses the debate about the value of pruning~\cite{crowley2018closer,liu2019rethinking}. Given the limited length, we defer the detailed trivial experimental settings to supplementary material. Important settings (\eg, LR) will be explicitly emphasized.

\subsection{Analysis with MLP-7-Linear} \label{subsec:analysis_mlp}
In a nutshell, in this section, we investigate how finetuning affects dynamical isometry (measured by mean JSV) and how LR plays a part in it.

\vspace{0.500em}
\noindent \bb{Evaluated network}. The network here is a 7-layer MLP (multi-layer perceptron) \emph{without non-linearity}. We adopt this network following~\cite{lee2020signal}. We are aware that this toy network has little practical meaning, but it is very appropriate here for two reasons. First, in our analysis, we will need a method to recover DI \emph{exactly}. Up to date, this can \emph{only} be achieved on linear networks~(see Tab.~\ref{tab:jsv_different_nets}), to our best knowledge. Second, it is free from the intervention of modern CNN features (\eg, BN~\cite{ioffe2015batch}, residual~\cite{resnet}). By our observation, these will make things complex and prevent us from seeing \emph{consistent} results at this early analysis stage.

\vspace{0.500em}
\noindent \bb{LR schedule setup}. When we set different LR schedules (with different initial LRs), we will keep (1) the total number of epochs is the same, (2) the last LR is the same. We adopt the step LR schedule because of its broad use (the conclusions can equally generalize to other schedules like cyclical LR~\cite{smith2017cyclical} or LR with warm restarts~\cite{loshchilov2017sgdr}).

\begin{figure*}[t]
\centering
\begin{tabular}{c@{\hspace{0.005\linewidth}}c@{\hspace{0.005\linewidth}}c@{\hspace{0.005\linewidth}}c@{\hspace{0.005\linewidth}}c@{\hspace{0.005\linewidth}}c}
  \includegraphics[width=0.245\linewidth]{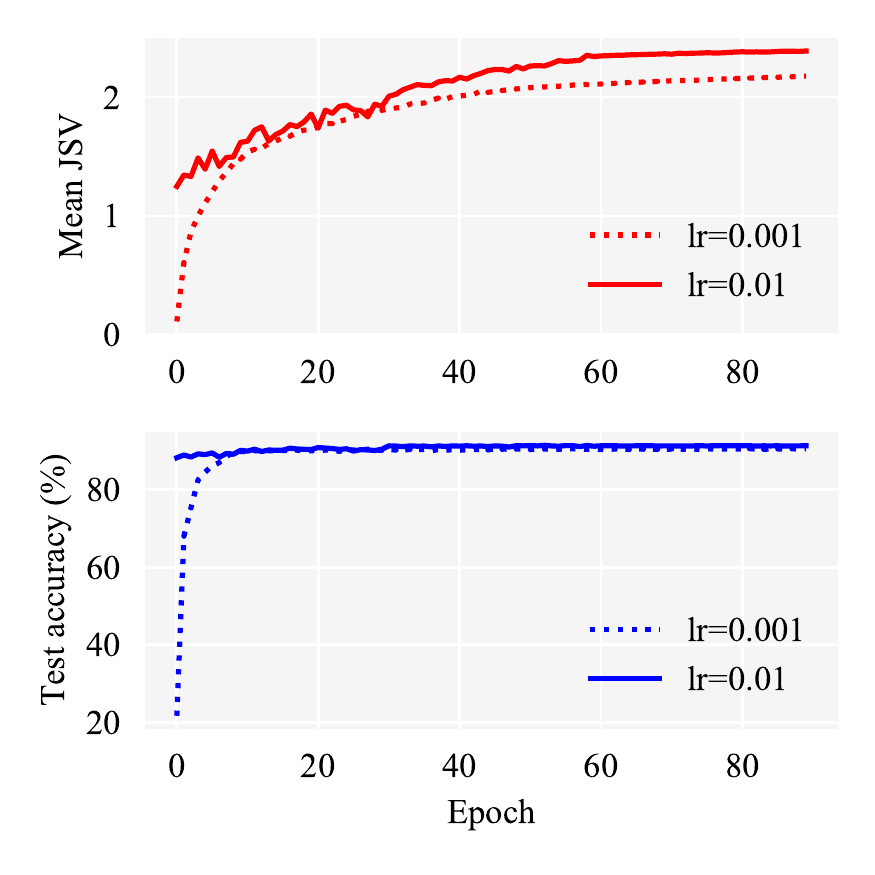} &
  \includegraphics[width=0.245\linewidth]{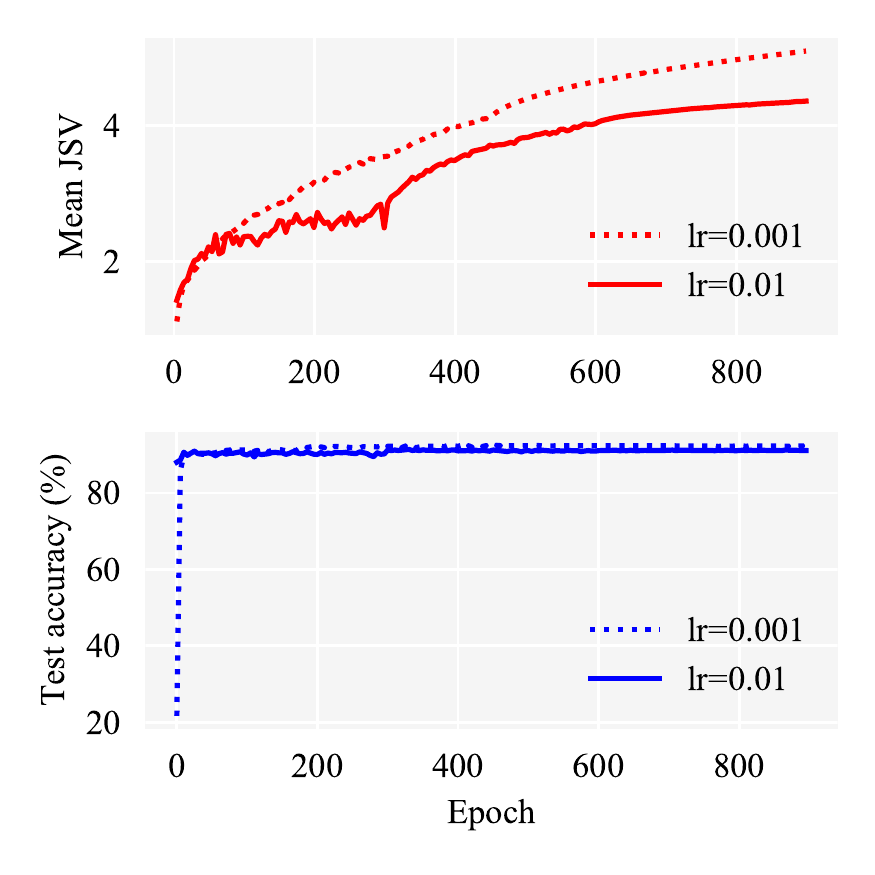} &
  \includegraphics[width=0.245\linewidth]{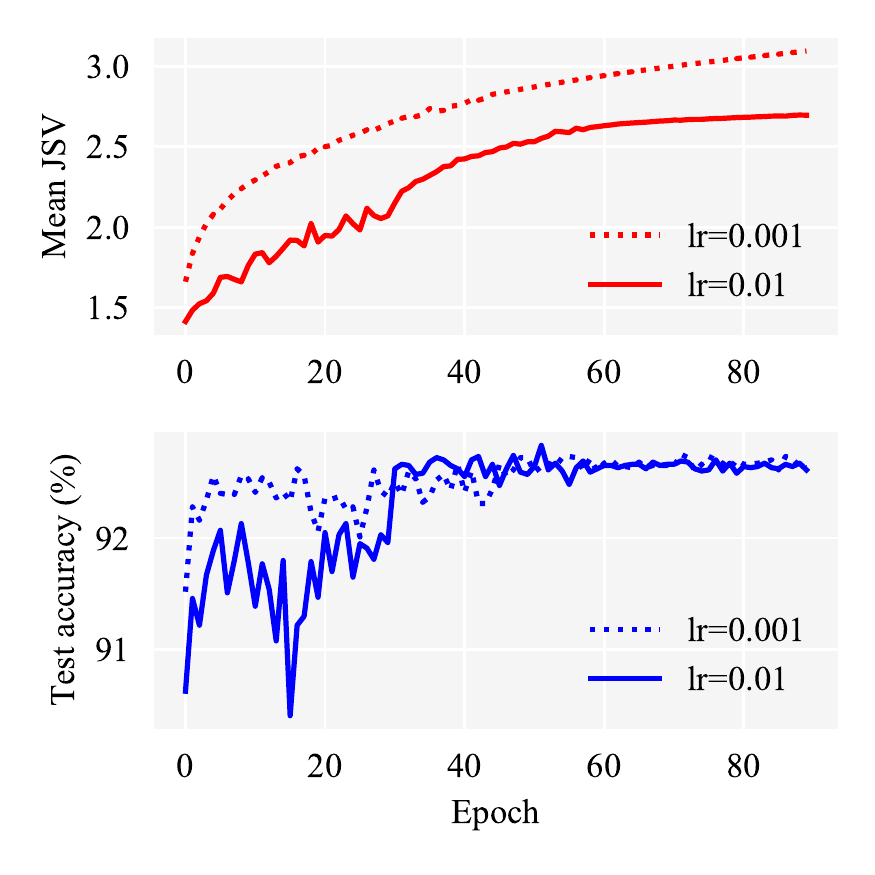} &
  \includegraphics[width=0.245\linewidth]{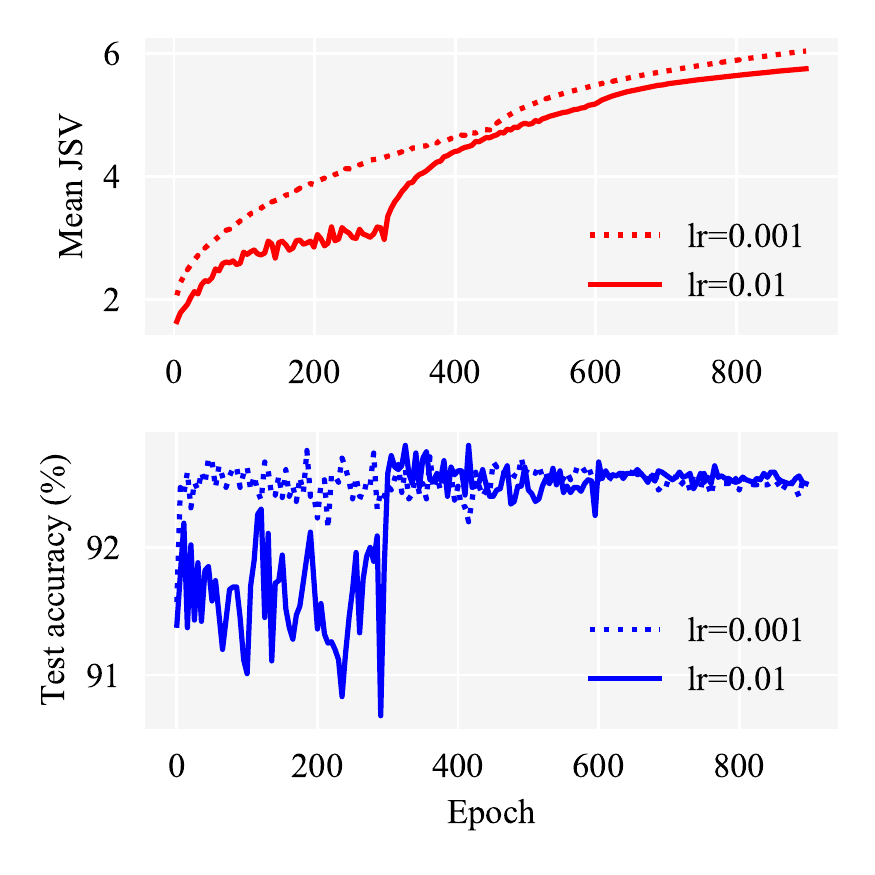} \\
  (a) 90 epochs & (b) 900 epochs  & (c) OrthP, 90 epochs & (d) OrthP, 900 epochs \\
\end{tabular}
\vspace{0.5000em}
\caption{Mean JSV and test accuracy during finetuning with different setups. Note that with OrthP (c, d), mean JSV recovers faster; so does the test accuracy. The pruning ratio in this case is 0.8. Please refer to the supplementary material for similar plots of pruning ratio 0.9.}
\label{fig:acc_jsv_plotts}
\vspace{-1em}
\end{figure*}

\vspace{0.500em}
\noindent \bb{Proposed hypotheses}.
We first plot the mean JSV in the finetuning process of the pruned MLP-7-Linear (pruning ratio 0.8). To our surprise, without any extra help, the mean JSV can recover itself during finetuning (see Fig.~\ref{fig:acc_jsv_plotts}, Row 1). Aware of this, it is straightforward to conjecture that any factor (\eg, learning rate) involved in the training can affect dynamical isometry. Concretely, we propose the following plausible explanation to the effect of a larger LR in finetuning~\cite{renda2020comparing,le2021network}:  \emph{A larger LR helps the network converge faster, thus the dynamical isometry (measured by mean JSV) recovers faster. Since better dynamical isometry promotes faster convergence, a larger LR leads to better final performance.} That is,  a larger LR in finetuning shows performance advantage because DI is damaged by pruning first. If DI is \emph{recovered} before finetuning, a larger LR probably does not have much advantage anymore.

The validation of the above explanation can be specified into the following 4 hypotheses:
\begin{itemize}
    \item \bb{Hypothesis 1}: Given a small number of epochs, mean JSV \emph{cannot} be fully recovered by training, then the large LR shows a significant advantage over small LR.
    \item \bb{Hypothesis 2}: Given an abundant number of epochs, mean JSV can be fully recovered by training, then the large LR has less advantage over the small LR.
    \item \bb{Hypothesis 3}: If we employ OrthP to recover the mean JSV, given the small number of epochs again, the large LR should have less advantage over the small LR.
    \item \bb{Hypothesis 4}: If we combine abundant epochs with OrthP, mean JSV will be recovered even completely, then the performance advantage of a large LR over a small LR will be even smaller.
\end{itemize}

Corresponding to these four hypotheses, the eight finetuning LR settings are summarized in Tab.~\ref{tab:mlp_lr_schedule}. The unpruned MLP model is trained with LR schedule ``90 epochs, 0:0.01,30:0.001,60:0.0001''. For pruning, we employ $L_1$-norm pruning~\cite{li2017pruning} throughout this paper because of its simplicity and prevailing use. Specifically, it sorts the neurons (or filters) by their $L_1$ norms in ascending order and prunes those with the least norms by a predefined pruning ratio $r$.

\begin{table}[t]
\centering  
\setlength\tabcolsep{3.5pt}
\begin{tabular}{lccc}
\toprule
Finetuning setting                   & LR 0.001  & LR 0.01      & Acc.~gain \\
\midrule
\multicolumn{4}{c}{Pruning ratio = 80\%} \\
90 epochs                   & 90.54 (0.02) &       \bb{91.36} (0.02)        & 0.82 \\
900 epochs                  & \bb{92.54} (0.03) &       91.64 (0.41)        & -0.90 \\
OrthP, 90 epochs            & 92.77 (0.03)   &       \bb{92.79} (0.03)      & 0.02 \\
OrthP, 900 epochs           & \bb{92.84} (0.03) &       92.81 (0.04)        & -0.03 \\
\hdashline 
Scratch                     &       \multicolumn{2}{c}{92.76 (0.03)} \\
\hline
\multicolumn{4}{c}{Pruning ratio = 90\%} \\
90 epochs                   & 87.59 (0.01)      &       \bb{87.81} (0.03)    & 0.22 \\
900 epochs                  & \bb{90.44} (0.01) &       87.83 (0.04)        & -2.61 \\
OrthP, 90 epochs            & 92.72 (0.03)      &       \bb{92.77} (0.04)   & 0.05 \\
OrthP, 900 epochs           & \bb{92.86} (0.03) &       92.79 (0.03)        & -0.07 \\
\hdashline 
Scratch                     &       \multicolumn{2}{c}{92.76 (0.04)} \\
\bottomrule
\end{tabular}
\vspace{0.5000em}
\caption{Test accuracies (\%) of the corresponding 4 hypotheses in Tab.~\ref{tab:mlp_lr_schedule}. Network: \bb{MLP-7-Linear}. Dataset: \bb{MNIST}. The accuracy of unpruned model is 92.77\%. Each setting is randomly run 5 times, mean (std) accuracy reported. ``Scratch'' stands for training the pruned model from scratch. ``Acc.~gain'' refers to the mean accuracy improvement of initial LR 0.01 over 0.001.}
\label{tab:mlp_7_linear_accuracy}
\vspace{-1em}
\end{table}

The final accuracy results are shown in Tab.~\ref{tab:mlp_7_linear_accuracy}, training plots shown in Fig.~\ref{fig:acc_jsv_plotts}. We first analyze the results of pruning ratio 0.8 in Tab.~\ref{tab:mlp_7_linear_accuracy}. As seen, when finetuned for 90 epochs, LR 0.01 shows an advantage over LR 0.001 by 0.82\% accuracy. It is tempting to draw a conclusion based on this comparison that LR 0.01 is much better than LR 0.001, as~\cite{renda2020comparing,le2021network} have found. However, this is not the whole story:
\begin{itemize}
    \item With 900 epochs, LR 0.01 is greatly surpassed by LR 0.001 (91.64 vs.~92.54). The reason is that, with abundant epochs, the dynamical isometry can be recovered more completely, hence LR 0.01 does not show advantage anymore over LR 0.001\footnote{As for the fact that LR 0.01 is largely \emph{surpassed} by LR 0.001, we cannot draw any generic conclusion about it, because on other experiments (Tab.~\ref{tab:other_toy_model_accuracy}) we do \emph{not} consistently observe a similarly strong performance of LR 0.001. One possible reason for this specific case is that, a small LR (\ie, 0.001) helps the network inherit more knowledge while LR 0.01 destroys most of the inherited knowledge.}.
    \item When OrthP applied, LR 0.01 does not show significant advantage either, similar to the effect of increasing the number of training epochs. This is because that finetuning shares the same role of recovering dynamical isometry with OrthP. Just OrthP is more effective since it is analytically targeting dynamical isometry.
    \item When the best setting used (OrthP + 900 epochs), LR 0.001 is slightly better than 0.01. Comparing ``OrthP, 900 epochs'' with ``OrthP, 90 epochs'', the gains are marginal. This is because the dynamical isometry has already been fully recovered by OrthP, thus more training epochs do not show much value anymore.
\end{itemize}
A different pruning ratio 0.9 is also explored. Results of it are in line with those of pruning ratio 0.8 in Tab.~\ref{tab:mlp_7_linear_accuracy}. 

Of special note is how misleading the results can be if we are not aware of the effect of DI in pruning. The ``90 epochs'' finetuning setting appears nothing wrong (considering the original unpruned network is trained for 90 epochs, finetuning for 90 epochs is not improperly short). However, it leads us to the partial conclusion that a larger LR is better than a smaller LR.  Since the two rows with OrthP present significantly better results than the other two. Clearly, the comparisons between LR 0.01 and LR 0.001 under these settings are most compelling. In this sense, it is fair to say LR 0.001 is as good as LR 0.01 (if not better). 

Besides, another lesson from Tab.~\ref{tab:mlp_7_linear_accuracy} is that, dynamical isometry recovery before finetuning is rather important for either better final generalization ability (note ``OrthP, 90 epochs'' delivers significantly better results than ``900 epochs'') or faster convergence (note the initial test accuracy is much higher when OrthP is used).

Next, we discuss how the discoveries from Tab.~\ref{tab:mlp_7_linear_accuracy} can help us unveil another mystery in pruning, \ie, the debate about the value of pruning.

\begin{table*}[t]
\centering
\setlength\tabcolsep{5.5pt}
\begin{tabular}{l|c|ccccc}
\toprule
Implementation                                           & Unpruned (\%)            & Pruned model                  & Scratch (\%)              & Pruned-Finetuned (\%)    & Finetuning LR schedule \\
\midrule
\multirow{2}{*}{Original paper~\cite{li2017pruning}}    & \multirow{2}{*}{73.23}    & ResNet-34-A                   & (Not reported)            & 72.56             & 20 epochs, 0.001, fixed \\
                                                        &                           & ResNet-34-B                   & (Not reported)            & 72.17             & 20 epochs, 0.001, fixed \\
\hline
\multirow{2}{*}{Rethinking~\cite{liu2019rethinking}}    & \multirow{2}{*}{73.31}    & ResNet-34-A                   & \bb{73.03}$^\dagger$                 & 72.56             & 20 epochs, 0.001, fixed \\  
                                                        &                                       & ResNet-34-B                       & \bb{72.91}$^\dagger$                 & 72.29             & 20 epochs, 0.001, fixed \\
\hline
\multirow{4}{*}{Our impl.}                              & \multirow{4}{*}{73.23}             & \multirow{4}{*}{ResNet-34-A}  & \multirow{4}{*}{73.62}    & 72.91             & 20 epochs, 0.001, fixed  \\
                                                        &                                    &                               &                           & 72.94             & 90 epochs, 0.001, fixed \\
                                                        &                                    &                               &                           & \bb{73.88}        & 90 epochs, 0.001, decay \\
                                                        &                                    &                               &                           & \bb{73.88}        & 90 epochs, 0.01,  decay \\
\hdashline
\multirow{4}{*}{Our impl.}                              & \multirow{4}{*}{73.23}             & \multirow{4}{*}{ResNet-34-B}  & \multirow{4}{*}{73.33}    & 72.50             & 20 epochs, 0.001, fixed \\ 
                                                        &                                    &                               &                           & 72.58             & 90 epochs, 0.001, fixed \\
                                                        &                                    &                               &                           & 73.61             & 90 epochs, 0.001, decay \\
                                                        &                                    &                               &                           & \bb{73.67}        & 90 epochs, 0.01,  decay \\
\bottomrule
\end{tabular}
\vspace{0.5000em}
\caption{Top-1 accuracy comparison of different implementations of the $L_1$-norm pruning~\cite{li2017pruning}. Network: \bb{ResNet-34}. Dataset: \bb{ImageNet}. $^\dagger$Here we cite the best scratch-training results of \cite{liu2019rethinking} (\ie, Scratch-B). We adopt the torchvision models as the unpruned models following common practices. The main point here is that \cite{liu2019rethinking} draws the conclusion that scratch training is better than pruning \emph{because of an improper finetuning LR scheme}. With proper finetuning LR schemes (``90 epochs, 0.001,  decay'' or ``90 epochs, 0.01,  decay''), pruning is actually \emph{better} than scratch training. Please refer to Sec.~\ref{subsec:value_of_pruning} for detailed discussions.}
\label{tab:L1_resnet34_imagenet}
\end{table*}

\begin{table*}[t]
\centering
\setlength\tabcolsep{5pt}
\begin{tabular}{l|ccc|ccc}
\toprule
Network                                     & PR     & Params reduc. (\%)    & FLOPs reduc. (\%)     & Scratch (\%)              & Pruned-Finetuned-1 (\%)         & Pruned-Fintuned-2 (\%) \\
\midrule
\multirow{7}{*}{ResNet-18}                  & 0                 & 0                     & 0                     & 69.76$^\dagger$                 & /                         & / \\
                                            & 0.1               & 9.56                  & 9.58                  & 70.12                     & \ul{70.45}                & \bb{70.52} (+0.07) \\
                                            & 0.3               & 28.32                 & 28.18                 & 69.02                     & \ul{69.32}                & \bb{69.50} (+0.18) \\
                                            & 0.5               & 47.03                 & 46.20                 & 67.02                     & \ul{67.37}                & \bb{67.75} (+0.18) \\
                                            & 0.7               & 65.99                 & 64.93                 & \ul{64.07}                & 63.73                     & \bb{64.41} (+0.68) \\
                                            & 0.9               & 84.75                 & 83.52                 & \bb{56.93}                & \ul{53.55}                & 53.23 (-0.22) \\
                                            & 0.95              & 89.51                 & 88.03                 & \ul{44.86}                & 44.35                     & \bb{47.63} (+3.28) \\
                                            
\hline
\multirow{7}{*}{ResNet-34}                  & 0                 & 0                     & 0                     & 73.23$^\dagger$           & /                         & / \\
                                            & 0.1               & 9.84                  & 9.92                  & 73.51                     & \ul{73.92}                & \bb{74.09} (+0.17) \\
                                            & 0.3               & 29.15                 & 29.26                 & 72.59                     & \ul{73.12}                & \bb{73.42} (+0.30) \\
                                            & 0.5               & 48.41                 & 48.12                 & 71.25                     & \ul{71.66}                & \bb{71.82} (+0.16) \\
                                            & 0.7               & 67.95                 & 67.63                 & 68.81                     & \ul{68.82}                & \bb{69.28} (+0.46) \\
                                            & 0.9               & 87.26                 & 86.97                 & 58.96                     & \ul{60.33}                & \bb{61.28} (+0.95) \\
                                            & 0.95              & 92.16                 & 91.69                 & \ul{53.81}                & 52.90                     & \bb{55.37} (+2.47) \\
\bottomrule
\end{tabular}
\vspace{0.5000em}
\caption{Top-1 accuracy comparison between scratch training (``Scratch'') and $L_1$-norm pruning~\cite{li2017pruning}. Network: \bb{ResNet-18, ResNet-34}. Dataset: \bb{ImageNet}. ``PR'' stands for pruning ratio. $^\dagger$We adopt the official torchvision models as unpruned models. ``Finetuned-1'' and ``Finetuned-2'' are short for the two finetuning LR schedules (``Finetuned-1'': 90 epochs, 0:1e-2,30:1e-3,60:1e-4,75:1e-5; ``Finetuned-2'': 90 epochs, 0:1e-3,45:1e-4,60:1e-5). In the parentheses of the last column is the relative accuracy gain of ``Pruned-Finetuned-2'' against ``Pruned-Finetuned-1''. Best accuracies are in \bb{bold} and second best \ul{underlined}. Please refer to Sec.~\ref{subsec:value_of_pruning} for detailed discussions.}
\label{tab:resnet_imagenet_L1_pruning}
\vspace{-1em}
\end{table*}

\subsection{Rethinking \emph{\bb{Again}} the Value of Pruning}~\label{subsec:value_of_pruning}
As far as we know, mainly \emph{two} papers question the value of inheriting weights from a pretrained model:~\cite{crowley2018closer,liu2019rethinking}\footnote{Both papers initially appeared in the NeurIPS 2018 CNNRIA Workshop. Interested readers may refer to the informative open discussions for more details (https://openreview.net/forum?id=r1eLk2mKiX, https://openreview.net/forum?id=r1lbgwFj5m).}. Both papers draw two similar conclusions. (1) Pruning has \emph{no} value, \ie, training from scratch the small model can match (or outperform sometimes) the counterpart pruned from a big pretrained model. (2) Given the fact of (1), what really matters in pruning may lie in the pruned \emph{architecture} instead of the inherited weight values. As such, both papers propose to view pruning as a form of neural architecture search. To our best knowledge, the questioning still remains an open debate in the community. In this section, we show how our findings above can help end this debate. In a nutshell, we conclude the \emph{opposite} way to these two papers: finetuning a pruned model is better than training from scratch in structured pruning (filter pruning).

\begin{table*}[t]
\centering
\setlength\tabcolsep{2.5pt}
\begin{tabular}{l@{\hspace{0.0002\linewidth}}ccccccccccccc}
\toprule
\multirow{2}*{Method}                           & \multirow{2}*{Scratch?} & \multicolumn{2}{c}{500K params budget}      & & \multicolumn{2}{c}{1M params budget}            & & \multicolumn{2}{c}{1.5M params budget} \\
                                                                            \cline{3-4}                                 \cline{6-7}                                         \cline{9-10}
                                                &                           & Params (M)    & Acc.~(\%)                          & & Params (M)  & Acc.~(\%)                                  & & Params (M)  & Acc.~(\%) \\
\midrule
$L_1$-norm pruning~\cite{li2017pruning}         & \xmark    & 0.51              & 90.86                                 & & 1.02    & 92.61                         & & 1.52    & 93.63 \\
Fisher pruning~\cite{theis2018faster}           & \xmark    & 0.52              & 92.59                                 & & 1.02    & 93.51                         & & 1.52    & 94.51 \\
Varying Depth                                   & \cmark    & 0.69              & 93.56                                 & & 1.08    & \ul{94.54}                    & & 1.47    & 94.64 \\
Varying Width                                   & \cmark    & 0.50              & 93.45                                 & & 0.98    & 94.30                         & & 1.48    & 94.66 \\
Varying Bottleneck                              & \cmark    & 0.50              & \ul{93.69}                            & & 1.00    & 94.40                         & & 1.49    & \ul{94.79} \\
Fisher Scratch                                  & \cmark    & 0.52              & \bb{93.72}                            & & 1.02    & \bb{94.65}                    & & 1.52    & \bb{94.86} \\
\hdashline
$L_1$-norm pruning~\cite{li2017pruning} (Rerun)     & \xmark    & 0.50          & 91.23                                 & & 1.00    & 92.80                         & & 1.51    & 93.52 \\
$L_1$-norm pruning~\cite{li2017pruning} (LR 0.01)   & \xmark    & 0.50          & \ul{93.88} (0.10)                     & & 1.00    & \ul{94.49} (0.10)             & & 1.51    & \ul{94.92} (0.16) \\
Fisher pruning~\cite{theis2018faster} (Rerun)       & \xmark    & 0.52          & 92.17                                 & & 0.98    & 93.57                         & & 1.48    & 94.67 \\
Fisher pruning~\cite{theis2018faster} (LR 0.01)     & \xmark    & 0.52          & \bb{94.27} (0.09)                     & & 0.98    & \bb{94.80} (0.02)             & & 1.48    & \bb{95.10} (0.13) \\
\bottomrule
\end{tabular}
\vspace{0.5000em}
\caption{Test accuracy comparison between 2 pruning schemes and 4 scratch training schemes in~\cite{crowley2018closer}. Network: \bb{WRN-40-2} (unpruned accuracy: 95.08\%, params: 2.24M). Dataset: \bb{CIFAR-10}. Results above the dashline are directly cited from \cite{crowley2018closer}; results below the dashline are from our reproducing (with the official code of \cite{crowley2018closer} at https://github.com/BayesWatch/pytorch-prunes for fair comparison). ``Rerun'' means we rerun the code of \cite{crowley2018closer} as it is. ``LR 0.01'' means we redo the finetuing for the pruned models in ``Rerun'' \emph{using our finetuning LR schedule} (120 epochs, 0:1e-2,60:1e-3,90:1e-4). Finetuning is randomly repeated for 3 times, mean (std) accuracies reported. The main point here is that \cite{crowley2018closer} draws the conclusion that scratch training is better than pruning \emph{because of an improper finetuning LR scheme}. With the proper finetuning LR scheme, pruning is actually \emph{better} than scratch training. Please refer to Sec.~\ref{subsec:value_of_pruning} for detailed discussions.}
\label{tab:result_fisher}
\vspace{-1em}
\end{table*}

\vspace{0.500em}
\noindent \bb{Reexamination of \cite{liu2019rethinking}}.
Before presenting results, here are some important comparison setting changes worth our attention: (1) In~\cite{liu2019rethinking}, they compare training from scratch with \emph{six} pruning methods (five structured pruning methods~\cite{li2017pruning,luo2017thinet,liu2017learning,he2017channel,huang2018data} and one unstructured pruning method~\cite{han2015learning}). Here, we only focus on the \emph{$L_1$-norm pruning}~\cite{li2017pruning} on ImageNet. The main reason is that, $L_1$-norm pruning is a \emph{basic} method of prevailing use. If we can show it outperforms training from scratch already, it will be no surprise to see other more advanced pruning methods also outperform training from scratch. In this sense, $L_1$-norm pruning is the most representative method here for our investigation. (2) In \cite{liu2019rethinking}, they have two variants for the number of epochs in scratch training: ``Scratch-E'' and ''Scratch-B''. In the former, different small models are trained for a fixed number of epochs; in the latter, \emph{smaller} models are trained for \emph{more} epochs to maintain the same computation budget (Scratch-B is shown to be better in~\cite{liu2019rethinking}). Here, we use ``Scratch-E'' with changes: We train the model for abundant epochs (120 epochs) and decay LR to a very small amount (1e-5) to \emph{ensure the network is fully converged}, since accuracy comparison \emph{before} the final convergence does not really bear much meaning. 

Results are shown in Tab.~\ref{tab:L1_resnet34_imagenet}. In the implementations of~\cite{liu2019rethinking}, the pruned and finetuned model is outperformed by the scratch training one, hence their ``no value of pruning'' argument. We also reproduce their settings (the two rows of ``20 epochs, 0.001, fixed'' in ``Our impl.'') for confirmation of their argument. However, the finetuning LR schedule ``20 epochs, 0.001, fixed'' is actually sub-optimal -- using the proper ones (``90 epochs, 0.001, decay'' or ``90 epochs, 0.01, decay''), pruning \emph{outperforms} scratch training. 

Tab.~\ref{tab:L1_resnet34_imagenet} only presents two pruned models. To have a panorama view about the impact of finetuning LR \emph{over the full spectrum of pruning ratios}, we vary the pruning ratios from 0.1 to 0.95. Results are presented in Tab.~\ref{tab:resnet_imagenet_L1_pruning}. As seen, with a more proper finetune LR scheme (column ``Pruned-Fintuned-2'' vs.~``Pruned-Fintuned-1), the performance can be improved \emph{significantly}. In general, the larger the pruning ratio, the more of the improvement. Now, comparing the results of ``Pruned-Fintuned-2'' to those of ``Scratch'', we can see pruning outperforms scratch-training in most cases. Only one exception is PR 0.95 for ResNet-18\footnote{We repeated this experiment but little change was observed. For now, we do not have much clue about the reason so take it as an outlier.}. Despite it, we believe it is fair to claim pruning is better than scratch training, namely, \bb{pruning \emph{has} value}. \cite{liu2019rethinking} concludes oppositely because they faithfully re-implement $L_1$-norm pruning based on the settings described in the original paper of~\cite{li2017pruning}: fixed LR 1e-3, 20 epochs, which actually are \emph{far from optimal} as we know now.

\begin{table*}[t]
\centering  
\small
\setlength\tabcolsep{2.5pt}
\begin{tabular}{lccccccccccc}
\toprule
\multirow{2}{*}{Finetuning setting}                        & \multicolumn{3}{c}{MLP-7-ReLU (Unpruned: 98.16)}                                    &   & \multicolumn{3}{c}{LeNet-5-Linear (Unpruned: 98.64)}                  &   & \multicolumn{3}{c}{LeNet-5-ReLU (Unpruned: 99.16)} \\
                        \cline{2-4}                                                             \cline{6-8}                                                 \cline{10-12}
                         & {LR 0.01}  & {LR 0.001}  & Acc.~gain                             &   & {LR 0.01}  & {LR 0.001}  & Acc.~gain                  &   & {LR 0.01}  & {LR 0.001}  & Acc.~gain \\
\midrule
90 epochs               & \bb{94.76} (0.15)         & 93.21 (0.13)          & 1.55          &   & \bb{91.41} (0.01)     & 91.24 (0.03)      & 0.17      &   & \bb{94.26} (0.25)         & 93.70 (0.46)              & 0.56 \\
900 epochs              & \bb{94.93} (0.23)         & 94.25 (0.13)          & 0.68          &   & 91.57 (0.02)          & \bb{91.59} (0.02) & -0.02     &   & 95.37 (0.40)              & \bb{95.59} (0.75)         & -0.22 \\
OrthP, 90 epochs        & \bb{93.88} (0.24)         & 93.44 (0.31)          & 0.44          &   & \bb{91.35} (0.02)     & 91.02 (0.31)      & 0.33      &   & \bb{95.77} (0.49)         & 94.48 (0.08)              & 1.29 \\
OrthP, 900 epochs       & 94.28 (0.37)              & \bb{94.47} (0.21)     & -0.19         &   & 91.36 (0.02)          & \bb{91.44} (0.21) & -0.08     &   & 96.26 (0.41)              & \bb{96.46} (0.07)         & -0.20 \\
\hdashline 
Scratch                 & \multicolumn{3}{c}{93.72 (0.36)}                                  &   & \multicolumn{3}{c}{91.16 (0.18)}                      &   & \multicolumn{3}{c}{95.44 (0.32)} \\
\bottomrule
\end{tabular}
\vspace{0.5000em}
\caption{Test accuracies (\%) of the corresponding 4 hypotheses in Tab.~\ref{tab:mlp_lr_schedule}. Networks: \bb{MLP-7-ReLU, LeNet-5-Linear, LeNet-5-ReLU}. Dataset: \bb{MNIST}. The accuracy (\%) of each unpruned model is indicated beside the model name (Unpruned). The pruning ratio is 90\% for the first 6 fc layers of MLP-7 and for the first 3 conv layers of LeNet-5. Each setting is randomly run 5 times, mean (std) accuracy reported. ``Scratch'' stands for training the pruned model from scratch. ``Acc.~gain'' refers to the mean accuracy improvement of initial LR 0.01 over 0.001. In brief, this table is an extension of Tab.~\ref{tab:mlp_7_linear_accuracy}, from linear MLP to non-linear MLP and CNN.}
\label{tab:other_toy_model_accuracy}
\vspace{-1em}
\end{table*}

Here we discuss the reason that the larger the pruning ratio, the more of the accuracy improvement, observed in Tab.~\ref{tab:L1_resnet34_imagenet}. As we have shown above, a \emph{larger} LR helps recover dynamical isometry \emph{faster}. When the pruning ratio is \emph{larger}, dynamical isometry is damaged \emph{more}. Therefore, a larger LR is more helpful and finally results in a greater accuracy improvement. Ideally, as inspired by Tab.~\ref{tab:mlp_7_linear_accuracy}, if there is some method to recover the dynamical isometry, the advantage of LR 0.01 over LR 0.001 should be diminished.

\vspace{0.500em}
\noindent \bb{Reexamination of~\cite{crowley2018closer}}. Coincidentally,~\cite{crowley2018closer} adopts a \emph{very similar} finetuning LR scheme to~\cite{liu2019rethinking}: they finetune the pruned network with the \emph{lowest} LR during scratch training (8e-4, similar to 1e-3 in \cite{liu2019rethinking}) and \emph{fixed}. Therefore, we cannot help conjecturing that the results of \cite{crowley2018closer} do not present the whole picture, either. With a proper finetuning LR schedule, their conclusion probably does not hold anymore. To confirm this, we reproduce the experiments of \cite{crowley2018closer}. 

Results are shown in Tab.~\ref{tab:result_fisher}. The original finetuning LR scheme is 8e-4, fixed. We adopt a larger initial LR (1e-2) and decay it. Exactly the same as the case of \cite{liu2019rethinking} (Tab.~\ref{tab:L1_resnet34_imagenet}), the argument of ``no value of pruning'' (note the results above the dashline) is built upon a sub-optimal finetuning LR scheme. If the proper one is used (note the results below the dashline), pruning actually \emph{outperforms} the best scratch training scheme (namely, ``Fisher Scratch'') consistently across different parameter budgets.

\vspace{0.500em}
\noindent \bb{Retrospective remarks}. Simply put, results above show that the debate about the value of pruning is largely attributed to sup-optimal finetuning settings. It is worthwhile to ponder at this point why this simple reason was not spotted for years. In fact, the problem is not so straightforward to see, because it has been broadly believed that inherited weights of a pruned model already pose it at a location close to the final solution in the loss landscape, hence no need to finetune the model for \emph{many} epochs (typically much fewer than the number of epochs to train the model from scratch). This is probably why~\cite{li2017pruning} finetunes the model for merely 20 epochs on ImageNet, far from the best setting. The reality, however, turns out to be the opposite way: The pruned models actually demand \emph{more} finetuning epochs because pruning hurts the dynamical isometry, slowing down the convergence. This gap between our presumption and the reality, not noticed by previous works, finally leads to the debate about the value of pruning. Now, after discovering the role of dynamical isometry in pruning, our paper clears much of the mystery. The value of pruning is also justified.

\section{Dynamical Isometry Recovery (DIR)}
From the investigations above, we know dynamical isometry recovery is necessary for pruning. In this section, we evaluate the efficacy of the proposed OrthP on more complex networks than linear MLP. Especially, how non-linearity (\eg, ReLU~\cite{nair2010rectified}) and convolution operation affect its effectiveness is of interest. Therefore, we conduct more experiments similar to Tab.~\ref{tab:mlp_7_linear_accuracy} with three different types of networks: MLP-7-ReLU, LeNet-5-Linear, LeNet-5-ReLU.

Results are presented in Tab.~\ref{tab:other_toy_model_accuracy}. Compared to the linear MLP case (Tab.~\ref{tab:mlp_7_linear_accuracy}), key observations include (1) training for more epochs (900 vs.~90) is still beneficial regardless of the LR schemes (because more training help recover dynamical isometry as shown in Fig.~\ref{fig:acc_jsv_plotts}); (2) OrthP does not always help: comparing ``OrthP, 90 epochs'' to ``90 epochs'' or ``OrthP, 900 epochs'' to ``900 epochs'', OrthP only helps in the LeNet-5-ReLU case; (3) comparing 900 epochs to 90 epochs in the ``Acc.~gain'' column, we can \emph{consistently} see the advantage of LR 0.01 over 0.001 is diminished or even totally disappear occasionally (note the minus gains). 

The 1st and 3rd observations are consistent with the linear MLP case in Tab.~\ref{tab:mlp_7_linear_accuracy}, further justifying our hypotheses about the role of dynamical isometry in pruning. The 2nd observation suggests OrthP does not work well in the networks of more practical interest. This is not very surprising since OrthP is designed for linear MLPs. To have a more effective and efficient DIR method for modern practical networks is worth future exploration. Before we have it, the vanilla SGD training can serve our need for now given its implicit dynamical isometry recovery effect.

\section{Conclusion}
In this work, we are motivated by an interesting finding in pruning that using a larger finetuning LR can improve the final performance significantly. We examine the reason behind through the lens of dynamical isometry, two key phenomena observed: (1) pruning more parameters hurts dynamical isometry more; (2) finetuning can recover dynamical isometry and a larger LR helps recover it faster. The performance advantage of a larger LR arises as a result of the interaction among network type, pruning ratio, and the number of finetuning epochs, fundamentally driven by the two observations above. On top of this explanation, we reexamine the debate about the value of pruning. We demonstrate that the ``no value of pruning'' argument is built upon sup-optimal finetuning LR schemes; with the proper ones, pruning \emph{does} have value. Furthermore, we evaluate the effectiveness of dynamical isometry recovery on non-linear and convolutional networks, which is still far from satisfaction, thus can be a worthy direction of future works.

This work adds new knowledge in many ways. (1) It promotes a perspective shift of the prevailing three-step pruning pipeline: placing finetuning in the center and taking pruning as the initialization for finetuning. (2) It is the first work to marry dynamical isometry with conventional pruning (not pruning at initialization) and show it plays a critical role in unveiling many (at least two) mysteries about pruning. (3) It let us know dynamical isometry recovery is a rather necessary step in pruning. If it is not well cared for, the performance comparison between different pruning methods may be partial or even misleading. We hope this work can help the community towards a better understanding of pruning, better benchmarks of different pruning methods, and stronger pruning algorithms in practice.

{\small
\bibliographystyle{ieee_fullname}
\bibliography{references}
}

\end{document}